\documentclass[journal]{IEEEtran}

\usepackage{cite}

\usepackage{color}

\usepackage{amsmath}

\usepackage{algorithm}
\usepackage{algorithmic}
\usepackage{bm}
\usepackage{latexsym}
\usepackage{amsthm}
\usepackage{url}
\usepackage{amsfonts}
\usepackage{amssymb}
\usepackage{indentfirst}
\usepackage{float}

\ifCLASSINFOpdf
   \usepackage[pdftex]{graphicx}
   \graphicspath{{../pdf/}{../jpeg/}}
   \DeclareGraphicsExtensions{.pdf,.jpeg,.png}
\else
   \usepackage[dvips]{graphicx}
   \graphicspath{{../eps/}}
   \DeclareGraphicsExtensions{.eps}
\fi
\usepackage{array}
\usepackage{multirow}

\ifCLASSOPTIONcompsoc
  \usepackage[caption=false,font=normalsize,labelfont=sf,textfont=sf]{subfig}
\else
  \usepackage[caption=false,font=footnotesize]{subfig}
\fi

\usepackage{setspace}
\begin{document}

%
\title{ {A self-supervised CNN for image watermark removal}}
%
%
%

\author{Chunwei Tian, \emph{Member}, \emph{IEEE},
        Menghua Zheng,
        Tiancai Jiao, 
        Wangmeng Zuo, \emph{Senior Member}, \emph{IEEE},
        Yanning Zhang, \emph{Senior Member}, \emph{IEEE},
        Chia-Wen Lin, \emph{Fellow}, \emph{IEEE}
\thanks{This work was supported in part by the National Natural Science Foundation of China under Grant
62201468, in part by the China Postdoctoral Science Foundation under Grant 2022TQ0259 and 2022M722599.

(Corresponding author: Yanning Zhang (Email: ynzhang@nwpu.edu.cn), Chia-Wen Lin (Email: cwlin@ee.nthu.edu.tw)) }
\thanks{Chunwei Tian is with the School of Software, Northwestern Polytechnical University, Xi’an, Shaanxi, 710129, China. Also, he is with the National Engineering Laboratory for Integrated Aero-Space-Ground-Ocean Big Data Application Technology, Xi’an, Shaanxi, 710129, China. (Email: chunweitian@nwpu.edu.cn)}
\thanks{Menghua Zheng is with the School of Software, Northwestern Polytechnical University, Xi’an, Shaanxi, 
710129, China. (Email: menghuazheng@mail.nwpu.edu.cn)}
\thanks{Tiancai Jiao is with the School of Software, Northwestern Polytechnical University, Xi’an, Shaanxi, 
710129, China. (Email: jtc@mail.edu.nwpu.cn)}
\thanks{Wangmeng Zuo is with the School of Computer Science and Technology, Harbin Institute of Technology, Harbin, Heilongjiang, 150001, China. Also, he is with the Peng Cheng Laboratory Laboratory, Shenzhen, Guangdong, 518055, China. (Email: wmzuo@hit.edu.cn)}
\thanks{Yanning Zhang is with the School of Computer Science, Northwestern Polytechnical University, the National Engineering Laboratory for Integrated Aero-Space-Ground-Ocean Big Data Application Technology, Xi’an, Shaanxi, 710129, China. (Email:ynzhang@nwpu.edu.cn)}
\thanks{Chia-Wen Lin is with the Department of Electrical Engineering and the Institute of Communications Engineering, National Tsing Hua University, Hsinchu300, Taiwan (E-mail: cwlin@ee.nthu.edu.tw)}
}


\maketitle

\begin{abstract}
Popular convolutional neural networks mainly use paired images in a supervised way for image watermark removal. However, watermarked images do not have reference images in the real world, which results in poor robustness of image watermark removal techniques. In this paper, we propose a self-supervised convolutional neural network (CNN) in image watermark removal (SWCNN). SWCNN uses a self-supervised way to construct reference watermarked images rather than given paired training samples, according to watermark distribution. A heterogeneous U-Net architecture is used to extract more complementary structural information via simple components for image watermark removal. Taking into account texture information, a mixed loss is exploited to improve visual effects of image watermark removal. Besides, a watermark dataset is conducted. Experimental results show that the proposed SWCNN is superior to popular CNNs in image watermark removal. Codes can be obtained at https://github.com/hellloxiaotian/SWCNN.

\end{abstract}


\section{Introduction}
Due to developments of big data and internet techniques, images have become a medium of office and entertainment. To protect copyrights of these images, watermarks (i.e., letter, number and logos) are conducted on given images to protect material \cite{braudaway1997protecting, liu2016blind}. Although these watermarks can claim ownerships of protected images under certain conditions, they are faced with some challenges of robustness of watermark techniques and network security \cite{lee2001survey}. To test quality of these watermarks, image watermark attack methods are presented \cite{tsai2007color}.

 {Using multiple destruction, i.e., JPEG compression, low-pass filtering and noise pollution can attack watermarked images to verify robustness of a watermarking algorithm \cite{wong2003novel}. Alternatively, removing image watermarks is also a good choice to protect copyright of added watermarks and protected clean images \cite{hu2005algorithm}. For instance, exploiting correct user keys in the wavelet domain can effectively remove watermarks to obtain clean images to protect copyright of users \cite{hu2005algorithm}. Combining user key and hidden data in the non-watermarked area can help users to remove embedded watermarks to obtain clean images\cite{hu2006reversible}.} Additionally, the cross-channel correlation from color images and structural information are used to  repair areas damaged by watermark \cite{park2012identigram}.  {Using locally dependent parts of given watermark images can automated remove watermarks \cite{westfeld2008regression}.} Taking into efficiency account, combining Canny edge operator, Otsu thresholding and total variation can deal with structure and texture images for enhancing attack ability  \cite{santoyo2017automatic}. Although these methods have obtained good effects in image watermark removal, they may suffer from drawbacks of manual tuning parameters and complex optimization parameter methods. 

To resolve these issues, deep networks with architectures of black boxes are presented in computer vision tasks \cite{cong2022global, cong2021rrnet}. For instance, Tian et al. used heterogeneous convolutions to extract horizontal and vertical key features to enhance visual effects in image super-resolution tasks \cite{tian2021asymmetric}. Due to strong learning ability, deep networks are also extended to image watermark removal \cite{chen2019leveraging}. A deep network with a fine-tuning operation can use fewer data to train a good image watermark removal model\cite{wang2020watermarking}. To test effectiveness of visible watermarks, a common framework based on a CNN is used to detect and remove watermarks \cite{cheng2018large}.  {To deal with watermark removal of realistic watermarked images, generative adversarial networks (GANs) are developed \cite{li2019towards}. Fusing a GAN and self-attention mechanism can remove watermarks of unknown locations to obtain high-quality images \cite{cao2019generative}. Besides, using elastic weight consolidation and unlabeled data augmentation to implement a fine-tuning framework can effectively deal with image watermark removal\cite{chen2021refit}.}  {To restore more texture details, a two-stage network containing a multi-task network and an attention is designed to recover more texture information rather than detecting watermark locations for watermark removal \cite{cun2021split}. Due to a barrier of uncertainty, i.e., size, shape and color, a two-phase network is presented \cite{liu2021wdnet}. That is, the first phase is used to obtain a rough decomposition via a given watermarked image. The second phase is exploited to remove watermark via centers on a watermark area. To extract robust watermark features, a cascaded U-Net network is conducted for watermark removal \cite {fu2022improved}. } Although these methods have obtained excellent performance for verifying robustness of image watermark removal, most of these methods require ground truth to train image watermark removal models, which is limited by real digital camera devices.
   
In this paper, we present a self-supervised convolutional neural network for image watermark removal as well as SWCNN. SWCNN utilizes a self-supervised mechanism to obtain reference watermarked images to test quality of added watermarks. Also, a heterogeneous U-Net architecture uses simple components to extract more complementary structural information for image watermark removal. To further improve quality of predicted images, a mixed loss is exploited to counterpoise structural information and texture information. Besides, a watermark dataset is conducted. The proposed SWCNN has obtained excellent results to verify watermark quality.

Contributions of the proposed SWCNN can be summarized as follows. 

 (1) A self-supervised mechanism is proposed to construct reference watermarked images rather than giving paired watermarked images. 
  
 (2)  {A heterogeneous U-Net with simple components is designed to obtain more hierarchical and complementary structural information in image watermark removal.} 
  
 (3)   {A mixed loss is utilized to extract more structural information and texture information for improving robustness of the proposed image watermark removal method.}
  
 (4)  {A watermark dataset is conducted via twelve novel watermarks, which is very convenient to engineers and scholars for commercial and civil.}  
 
Remaining parts of this paper can be summarized as follows. 
Section II gives proposed method. Section III illustrates experiments. Section IV describes conclusion.

\section{Proposed method}
 {\subsection{Self-supervised Mechanism}}
It is known that most of existing image watermark removal methods depend on paired watermarked image and non-watermarked image to verify their robustness  \cite{song2010analysis}. However, real watermarked images do not have reference clean images, which limits their applications on real digital camera devices.  {That can be address by machine learning \cite{lehtinen2018noise2noise}. That is, we assume that we test temperature of a fixed room many times to obtain unreliable results, i.e., $y={y_1,y_2,...}$. To obtain true temperature, a common method uses a discriminant function (also treated as loss function) $D$ to obtain a smallest average deviation to find a number $g$  \cite{lehtinen2018noise2noise}, which can be formulated as Eq. (1).}

\vspace{-0.2cm}
 {
\begin{footnotesize}
\begin{equation}
\mathop {\arg \min }\limits_g {E_y}\{ D\{ (g,y)\}, 
\end{equation}
\end{footnotesize}}
\vspace{-0.3cm}

 {When D is a L2 loss function and $D(g,y) = {(g - y)^2}$, its optimization solution can be obtained via  $g = {E_y}(y)$. When D is a L1 loss function and $D(g,y) = \left| {g - y} \right|$, its optimization solution can be obtained by the $g = median\{ y\}$. According to a statistical idea, common estimation can convert a loss function as a negative log likelihood function \cite{lehtinen2018noise2noise}. Also, training of a neural network is suitable to mentioned principle. Its training procedure can be expressed as Eq. (2). }

\vspace{-0.2cm}
 {\begin{footnotesize}
\begin{equation}
\mathop {\arg \min }\limits_\theta  {E_{(i,o)}}(D({f_\theta }(i),o)),
\end{equation}
\end{footnotesize}}
\vspace{-0.4cm}

\noindent  {where $i$ and $o$ are an input and a target, respectively. $\theta$ denotes parameters. Also, $f$ stands for a network function.}  

 {According to Bayesian theory, Eq. (2) can be replaced with Eq. (3) \cite{lehtinen2018noise2noise}. 
\begin{footnotesize}
\begin{equation}
\mathop {\arg \min }\limits_\theta  {E_i}\{ {E_{o\left| i \right.}}\{ D\{ {f_\theta }(i),o\} \} \} 
\end{equation}
\end{footnotesize}}
\vspace{-0.4cm}

 {It is known that input-conditioned target distributions can be transformed to arbitrary distributions when they have same conditional expected values. Thus, Eq. (3) can be further optimized as Eq. (4) \cite{lehtinen2018noise2noise}.}

\vspace{-0.1cm}
 {\begin{footnotesize}
\begin{equation}
\mathop {\arg \min }\limits_\theta  \sum\limits_j {D({f_\theta }(i_j^1),o_j^1)},
\end{equation}
\end{footnotesize}}
\vspace{-0.3cm}

\noindent  {where predicted value and target have noisy distribution and they are suitable to $E\{ o_j^1|i_j^1\}  = {o_j}$. Thus, different noisy images with same distribution noise can be conducted as paired noisy image dataset. Inspired by noise-to-noise idea  \cite{lehtinen2018noise2noise}, we propose a self-supervised mechanism to construct paired data of a watermarked image and reference image. That is, watermarks randomly added once on a given clean image is considered as a watermarked image and watermarks randomly added another is treated as a reference image to construct a paired of images between a watermarked image and reference image to conduct a training dataset where watermarked images are conducted via the mentioned illustrations. Also, watermarked images are conducted via the following equation  \cite{dekel2017effectiveness}.} 

\vspace{-0.1cm}
 {\begin{footnotesize}
\begin{equation}
 {I_w}(pi) = \alpha (pi)W(pi) + (1 - \alpha (pi)){I_c},
 \end{equation}
\end{footnotesize}}
 \vspace{-0.4cm}
 
\noindent  {where $pi(x,y)$ is a pixel location. $\alpha (pi)$ denotes spatially varying opacity. $I_w$ stands for a watermarked image and $W$ is used to represent a watermark. $I_c$ is utilized to symbol a natural image. More information can be found in Ref. \cite{dekel2017effectiveness}. }
 {\subsection{Network Architecture of Self-supervised Convolutional Neural Network}}
\begin{figure*}[!htbp]
\centering
\subfloat{\includegraphics[width=7in]{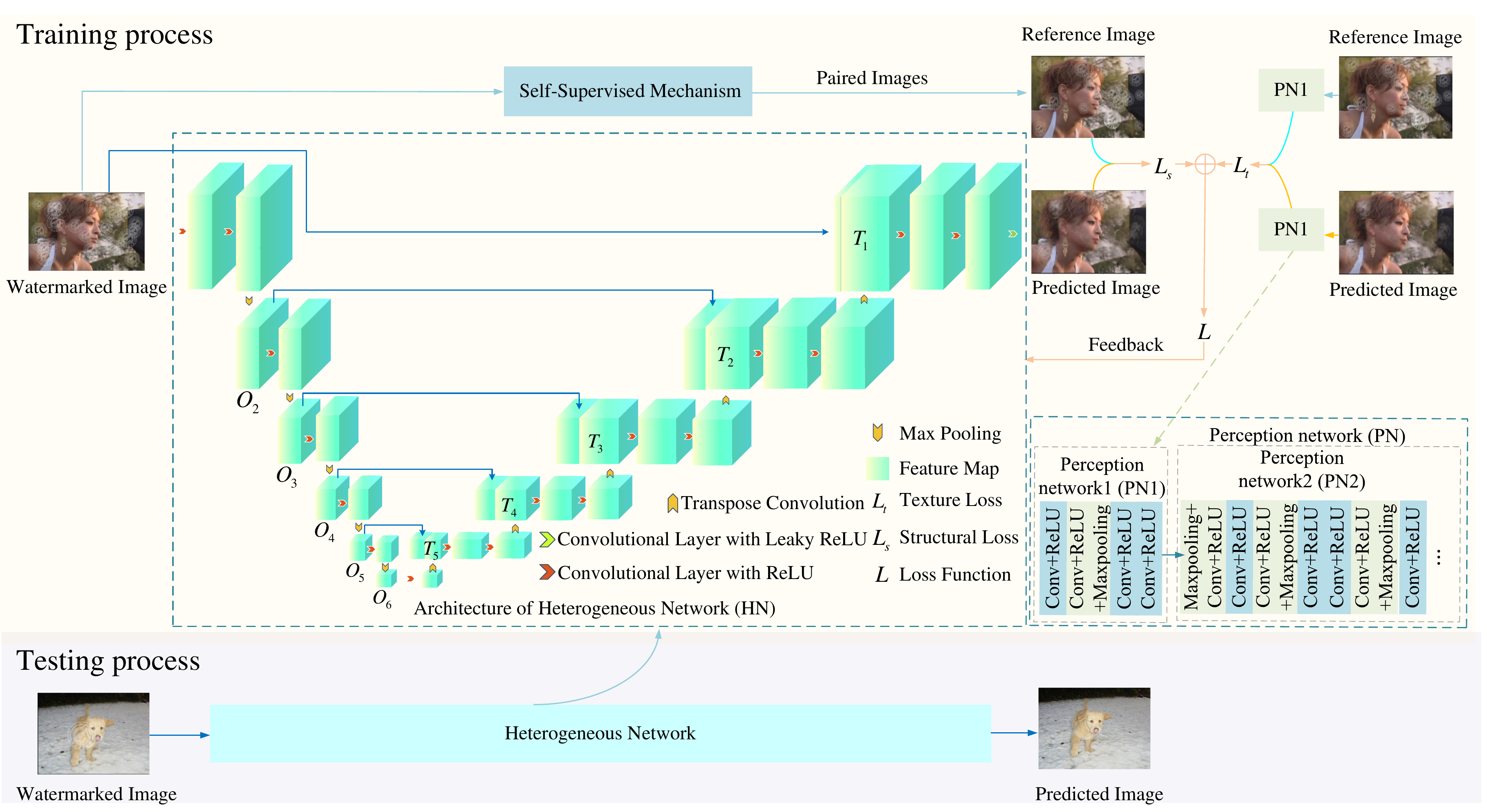}
  }
\caption{ \textcolor{black}{Architecture of SWCNN.}}
 
\end{figure*}

\linespread{1.2}
The SWCNN contains three parts, i.e., a self-supervised mechanism (SSM), an 18-layer heterogeneous network (HN) and a perception network (PN). The SSM uses noise-to-noise  \cite{lehtinen2018noise2noise} to construct reference watermarked images. The HN containing different activation functions, pooling functions, convolutional layers, concatenation operations and transpose convolutions uses obtained paired watermarked images and reference watermarked images from SSM to achieve heterogeneous U-Net architecture for facilitating more complementary structural information in image watermark removal. To obtain more texture information, a PN uses results of middle layers of trained a classifier by the ImageNet  \cite{krizhevsky2012imagenet} as well as PN1 in Fig. 1 to detect quality of obtained texture information. To visually express the mentioned process of SWCNN, we define the following equation.

\vspace{-0.3cm}
\begin{footnotesize}
\begin{equation}
\begin{array}{ll}
{O_{SWCNN}} &= {f_{SWCNN}}({I_w})\\
{\rm{   }} &= f_{PN1}({f_{HN}}({I_w})),
\end{array}
\end{equation}
\end{footnotesize}
\vspace{-0.5cm}

\noindent  {where} ${I_w}$ denotes given watermarked images, ${f_{SWCNN}}$ is function of SWCNN,  {$f_{PN1}$ and ${f_{HN}}$} represents functions of PN1 and HN. Also, ${O_{SWCNN}}$ is utilized to express the output of SWCNN. ${f_{SWCNN}}$ is obtained by the following loss function and training data from the SSM as shown Section II.E.

 {\subsection{Heterogenous Network}}
18-layer heterogeneous network as well as HN  \cite{lehtinen2018noise2noise} containing some components, i.e., convolutional layers, ReLU  \cite{krizhevsky2012imagenet}, Leaky ReLU\cite{maas2013rectifier}, max pooling operations, concatenation operations and transpose convolutions is used to extract different features for improving facilitating more complementary structural information in image watermark. The 1st, 8th, 10th, 12th, 14th, 16th and 17th layers are composed of a combination of a convolutional layer and ReLU. The 2nd, 3rd, 4th, 5th and 6th layers consist of a combination of a convolutional layer, ReLU and max pooling. The 7th, 9th, 11th, 13th and 15th layers include a combination of a convolutional layer, ReLU and transpose convolution. The 18th layer contains a combination of a convolutional layer and Leaky ReLU. To improve memory ability of the HN, concatenation operations are acted on different layers. That is, given watermarked images and obtained information of the 15th layer are concatenated as input of the 16th layer in the HN. Obtained features of the 2nd and 13th layers are fused via a concatenation operation to act the 14th layer in the HN. Obtained information of the 3rd and 11th layers are integrated via a concatenation operation to act the 12th layer in the HN. Obtained information of the 4th and 9th layers are integrated via a concatenation operation to act the 10th layer in the HN. Obtained information of the 5th and 7th layers are integrated via a concatenation operation to act the 8th layer in the HN. Kernel sizes of all the layers are  $3 \times 3$. The 1st layer has input channels of 3 and output channels of 48. Input and output channel number of the 2nd-7th layers are 48, respectively. The 8th and 9th layer have input channels of 96 and output channels of 96. The 10th, 12th and 14th layer have input channels of 144 and output channels of 96. The 11th, 13th and 15th layer have input channels of 96 and output channels of 96. The 16th layer has input channels of 99 and output channels of 64. The 17th layer has input channels of 64 and output channels of 32. The 18th layer has input channels of 32 and 3. The mentioned illustrations can be formulated as follows.


\textcolor{black}{
\begin{equation}
\begin{footnotesize}
\begin{aligned}
O_{HN} &= f_{HN}({I_w}) \\
&=CLR(CR(CR(Cat({I_w},{T_1}))))
\end{aligned}
\end{footnotesize}
\end{equation}
\vspace{-0.9cm}}

\textcolor{black}{
\begin{footnotesize}
\begin{equation}
 {{T_i} = CRTC(CR(Cat(O_{i+1},T_{i+1})))(i=1,2,3,4)}
\end{equation}
\end{footnotesize}
\vspace{-0.9cm}}

\textcolor{black}{
\vspace{-0.2cm}
\begin{footnotesize}
\begin{equation}
 {T_5} = CRTC(O_6)
\end{equation}
\end{footnotesize}
\vspace{-0.9cm}}

\textcolor{black}{
\vspace{-0.2cm}
\begin{footnotesize}
\begin{equation}
 {{O_2} = CRMP(CR({I_w})),}
\end{equation}
\end{footnotesize}
\vspace{-0.9cm}}

\textcolor{black}{
\vspace{-0.2cm}
\begin{footnotesize}
\begin{equation}
 {{O_3} = 2CRMP(CR({I_w}),}
\end{equation}
\end{footnotesize}
\vspace{-0.9cm}}

\textcolor{black}{
\vspace{-0.2cm}
\begin{footnotesize}
\begin{equation}
 {{O_4} = 3CRMP(CR({I_w}))},
\end{equation}
\end{footnotesize}
\vspace{-0.9cm}}

\textcolor{black}{
\vspace{-0.2cm}
\begin{footnotesize}
\begin{equation}
 {{O_5} = 4CRMP(CR({I_w})),}
\end{equation}
\end{footnotesize}
\vspace{-0.9cm}}

\textcolor{black}{
\vspace{-0.2cm}
\begin{footnotesize}
\begin{equation}
 {{O_6} = 5CRMP(CR({I_w})),}
\end{equation}
\end{footnotesize}
\vspace{-0.9cm}}
\\

\noindent  {where $Cat$ is a concatenation operation, $CRTC$ is defined as a combination of $CR$ (also regarded as a combination of a convolutional layer and a ReLU), Transpose convolution. $CLR$ is the combination of a convolutional layer and a Leaky ReLU. $CRMP$ is utilized to represent a convolutional layer, a ReLU and a Maxpooling operation. $O_i$ expresses an output of the ith layer in the heterogeneous network, where i=2,3,4,5,6. $T_i$ expresses an output of the ith transpose convolutional layer in the heterogeneous network, where i=1,2,3,4,5. $iCRMP$ denotes i stacked CRMP, where i =2,3,4,5. And $O_{HN}$ stands for a predicted image of the heterogeneous network.}

 {\subsection{Perception Network}}
Perception network is implemented via a VGG architecture \cite{li2019towards}. That is, we use VGG on ImageNet to train a classifier. Subsequently, we put predicted results of HN and obtained reference image as inputs of PN (The first four layers of PN are composed of PN1), respectively. The outputs of 4th layer in the PN from mentioned different inputs are used to compute loss value of a texture loss. $f_{PN1}$ in Eq. (6) can be represented as Eq. (15).

\vspace{-0.35cm}
\begin{footnotesize}
\begin{equation}
\begin{array}{ll}
{f_{PN1}} = CR(CR(CRMP(CR({f_{vgg}})))),
\end{array}
\end{equation}
\end{footnotesize}
\vspace{-0.65cm}

\noindent  {where $f_{vgg}$ is denoted as a function of VGG, which can be shown in Ref. \cite{li2019towards}. Also, network architecture of PN1 can be visually shown in Fig.1.}
\subsection{Loss Function}
Taking into structural and texture information account, we propose a mixed loss based L1, which contains two parts, i.e., a structural loss and texture loss. The structural loss function is responsible for monitoring robustness of extracting structural features from a heterogeneous network. Also, the texture loss is used to monitor robustness of obtaining texture information from a perception network, where texture information is intermediate results of a classification network via the ImageNet. To improve training speed, mentioned loss functions are implemented by L1. Besides, training data is obtained by a self-supervised mechanism in Section II.C. To vividly express mentioned mixed loss, the following equation is conducted.

\vspace{-0.5cm}
{\begin{scriptsize}
\begin{equation}
\begin{array}{ll}
{L} &= {L_s} +  \lambda{L_t}\\
\rm{  } &= {\raise0.7ex\hbox{$1$} \!\mathord{\left/ {\vphantom {1 N}}\right.\kern-\nulldelimiterspace}
\!\lower0.7ex\hbox{$N$}}\sum\limits_{i = 1}^N {\left| {{f_{HN}}(I_w^i) - I_r^i} \right| +  \lambda \left| {{f_{PN1}}({f_{HN}}(I_w^i)) - {f_{PN1}}(I_r^i)} \right|},
\end{array}
\end{equation}
\end{scriptsize}}
\vspace{-0.4cm}

\noindent  {where} $L$, $L_t$ and $L_s$ are used to denote loss functions of the SWCNN, structural loss function and texture information, respectively. $\lambda$ is the adjustment coefficient for the texture information. $N$ is the number of watermarked images. $I^i_r$ is used to represent the $ith$ obtained reference image and $I^i_w$ is exploited to the $ith$ watermarked image, which can be introduced in Section II.A. Besides, the trained model relies on Adam \cite{kingma2014adam} to optimize parameters.

\linespread{1.2}
 {According to mentioned illustrations, we can see that the proposed mixed loss is composed of two parts, i.e., a structural loss and texture loss. The structural loss depends on HN, which is conducted as follows.}

\vspace{-0.35cm}
\begin{footnotesize}
\begin{equation}
\begin{array}{ll}
{L_s} &= 1/N\sum\limits_{i = 1}^N {\left| {{f_{HN}}(I_w^i) - I_r^i} \right|}  \\
{\rm{   }} &= 1/N\sum\limits_{i = 1}^N {\left| {O_{HN}^i - I_r^i} \right|} ,
\end{array}
\end{equation}
\end{footnotesize}
\vspace{-0.35cm}

\noindent  {where $O^i_{HN}$ is the HN output of the $ith$ image. The texture loss relies on PN, which can be shown in Section II. D.} 
\section{Experiments}
\subsection{Conducted datasets}
To make obtained watermark method more robust, we collect a training and test datasets based an image with more watermarks rather than an image with one watermark.

Training datasets: We choose 477 representative natural images with format of ‘.jpg’ from the PASCAL VOC 2012  \cite{everingham2015pascal}. Specifically, each watermarked image of training dataset randomly uses one of twelve watermarks in Fig. 2 with one transparency from 0.3, 0.5, 0.7 and 1.0, from 0 to 0.4 in coverage in a self-supervised way in Section II.A is conducted, where watermark size is set 0.5-1 times. To accelerate training speed, each watermark image is cropped as 3,111 patches with $256\times256$.

Test datasets: We choose 27 representative natural images with format of ‘.jpg’ from the PASCAL VOV 2012  \cite{everingham2015pascal}. Specifically, each watermarked image is conducted via using each watermark of twelve watermarks with fixed transparency, according to Ref.  \cite{dekel2017effectiveness}. The number of test watermarked images is 324 and each watermark size is 1.5 times.

\begin{figure}[!htbp]
\centering
\subfloat{\includegraphics[width=3.5in]{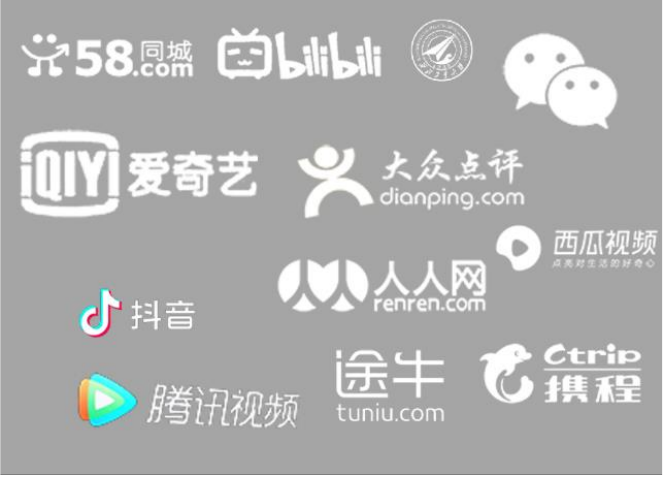}
  }
\caption{Twelve collected watermarks.}
 
\end{figure}

\subsection{Experimental setting}
All experiments are conducted on Ubuntu of 20.04 with Intel Xeon Silver 4210 CPU via PyTorch  \cite{paszke2017automatic} of 1.12 and Python of 3.8. To improve training speed, a GPU of 3090 with CUDA of 10.2 and CuDNN of 8.0.5 is used. Besides, batch size is 8, epoch number is 100 and initial learning rate is 1e-3, which may get 0.1 of original learning rate in 30th times. More parameters are the same as Ref.  \cite{zhang2017beyond}.
\subsection{Experimental analysis}
The proposed SWCNN has three key techniques, i.e., a self-supervised mechanism, 
a heterogenous architecture of HN and a mixed loss. Their effectiveness and rationality can be analyzed, according to training data, network design principle and image representation, respectively. 

A self-supervised mechanism: It is known that existing methods almost depend on paired data (i.e., watermarked images and clean images) in a supervised way to verify robustness of added watermarks in terms of training data\cite{cheng2018large}. However, corresponding clean images are difficultly obtained in real world. Inspired by Noise2Noise  \cite{lehtinen2018noise2noise}, we present a self-supervised mechanism to construct paired watermarked images, according to distributions of watermarks. That is, watermarks randomly added in a clean image once is regarded as a clean reference image as well as ground truth. Watermarks randomly added in a clean image another is regarded as a watermarked image as well as input of SWCNN. This can address drawback of no ground truth. Effectiveness of the proposed self-supervised mechanism is verified via both SWCNN and SWCNN with clean reference images in terms of PSNR and SSIM as shown in Table I. To fully verify performance of the proposed self-supervised mechanism, we monitor its performance for different steps of one epoch and different epochs. We can see that the proposed self-supervised mechanism performs better than given paired images, i.e., watermarked image and clean image in Fig.3. According to mentioned illustrations, we can see that the proposed self-supervised mechanism is very effective to remove watermarks.
\begin{table}[]
\caption{Removing watermark performance of different methods.}
\begin{tabular}{cccll}
\hline
Methods                             & PSNR    & SSIM       \\ \hline
 {HN with   clean reference images}   & 34.8979 & 0.9854     \\
HN                              & 31.6688 & 0.9803     \\
SWCNN                               & 36.9022 & 0.9893     \\
 {SWCNN with   clean reference images} & 35.8898 & 0.9881     \\ \hline

\end{tabular}
\end{table}
\begin{figure*}[!htbp]
\centering
\subfloat{\includegraphics[width=5.5in]{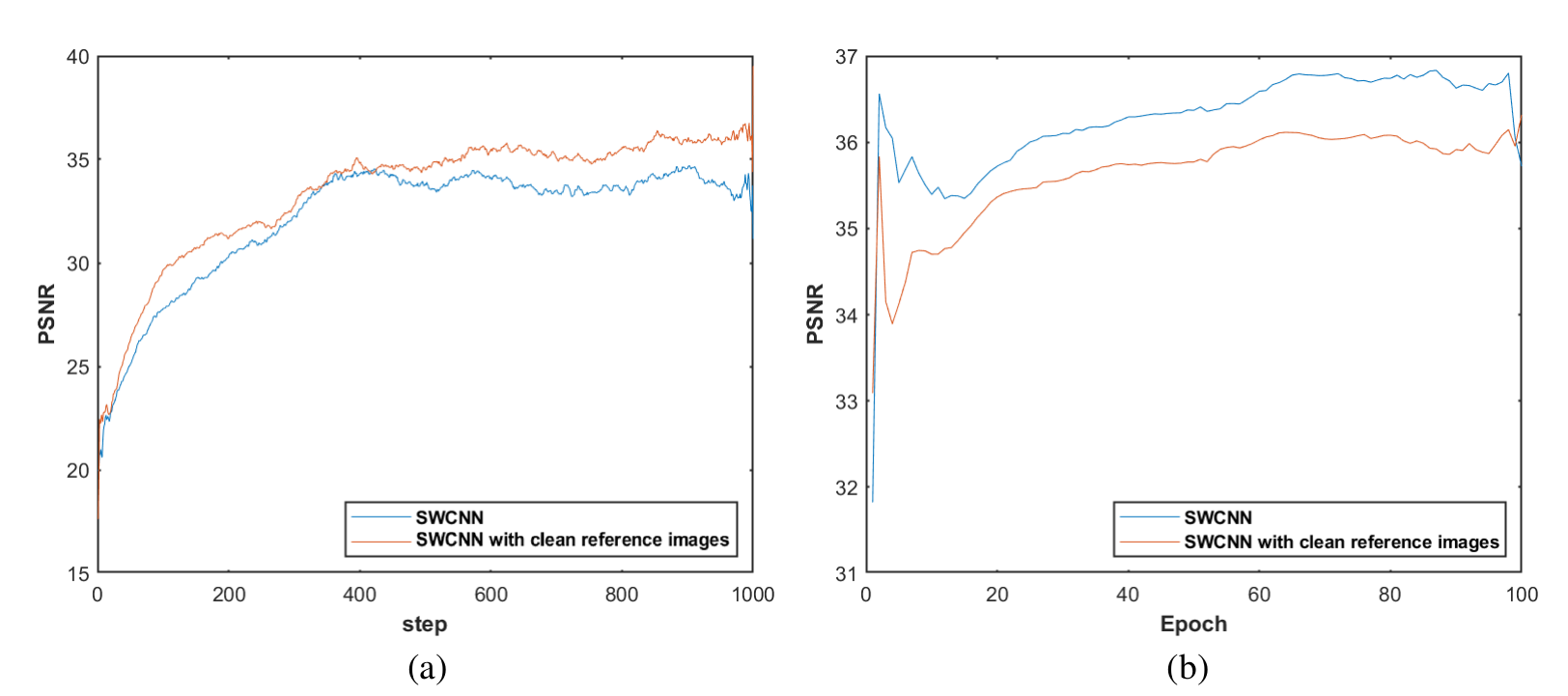}
  }
\caption{ {PSNR of different methods for one epoch and multiple epochs. (a)	PSNR of two methods for one epoch {,}  (b) PSNR of two methods for different epochs}}
\end{figure*}

Heterogenous network: It is known that expressive ability of deep networks is stronger, obtained structural information is more accurate, according to network design principle  \cite{tai2017memnet}. For instance, a U-Net combines different components, i.e., convolutional layer, ReLU, max pooling operations, concatenation operations and transpose convolutions to achieve a heterogenous network for extracting more structural information in image denoising  \cite{lehtinen2018noise2noise}. Motivated by that, we also use heterogenous network to mine complementary structural information to verifying quality of given watermarks. We use denoising CNN (DnCNN)  \cite{zhang2017beyond}, fast and flexible denoising CNN (FFDNet)  \cite{zhang2018ffdnet}, deep Image prior restoration (DIP)  \cite{ulyanov2018deep}, WGAN-GP  \cite{yu2018generative},  {detail-recovery image deraining network (DRD-Net) \cite{deng2019drd}, a deep residual learning algorithm for removing rain streaks (FastDerainNet) \cite{wang2020fastderainnet} and efficient attention fusion network in wavelet domain for demoireing (EAFNWDD) \cite{sun2021efficient}} as comparative methods on test dataset for transparency of 0.3 to test performance of the proposed SWCNN. As shown in Table II, effectiveness of heterogenous network is tested via comparing DnCNN, FFDNet, DIP and in terms of PSNR and SSIM for removing watermarks.
\begin{figure*}[!htbp]
\centering
\subfloat{\includegraphics[width=5.5in]{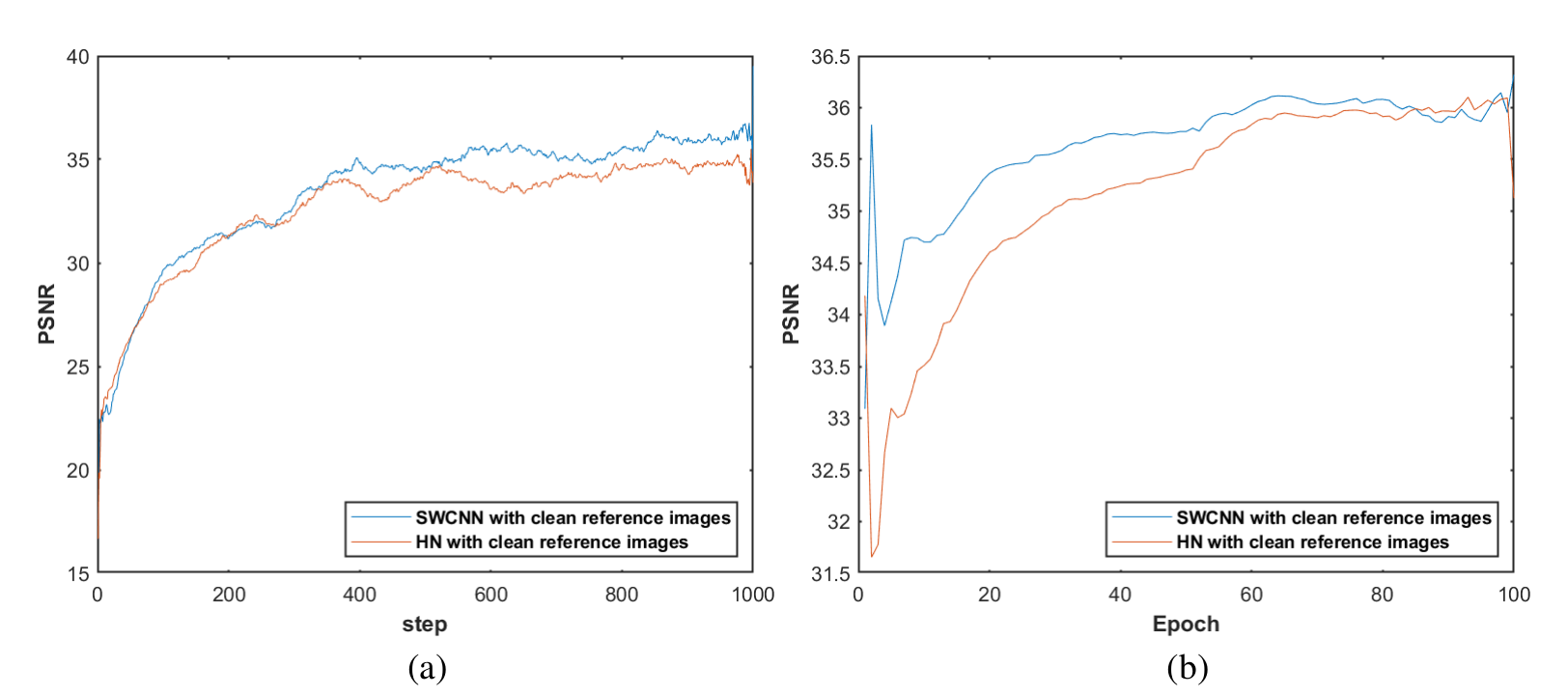}
  }
\caption{ {PSNR of different methods for one epoch and multiple epochs. (a) PSNR of two methods for one epoch, (b) PSNR of two methods for different epochs.}}
 
\end{figure*}
\begin{table}[]
\centering
\caption{Removing watermark performance of different methods for transparency of 0.3.}
\begin{tabular}{cccll}
\hline
Methods          & PSNR    & SSIM   \\ \hline
FFDNet  \cite{zhang2018ffdnet}  & 27.8820 & 0.8778 \\
DIP  \cite{ulyanov2018deep}     & 29.7473 & 0.9260 \\
WGAN-GP  \cite{yu2018generative} & 31.0752 & 0.9662 \\
DnCNN  \cite{zhang2017beyond}   & 30.1071 & 0.9620 \\
 {DRD-Net \cite{deng2019drd}}          & 28.9090 & 0.9707\\
 {FastDerainNet \cite{wang2020fastderainnet}}          & 32.2593 & 0.9815\\
 {EAFNWDD \cite{sun2021efficient}}          & 33.4744 & 0.9700\\
SWCNN (Ours)     & 36.9022 & 0.9893 \\ \hline
\end{tabular}
\end{table}

\begin{figure*}[!htbp]
\centering
\subfloat{\includegraphics[width=5.5in]{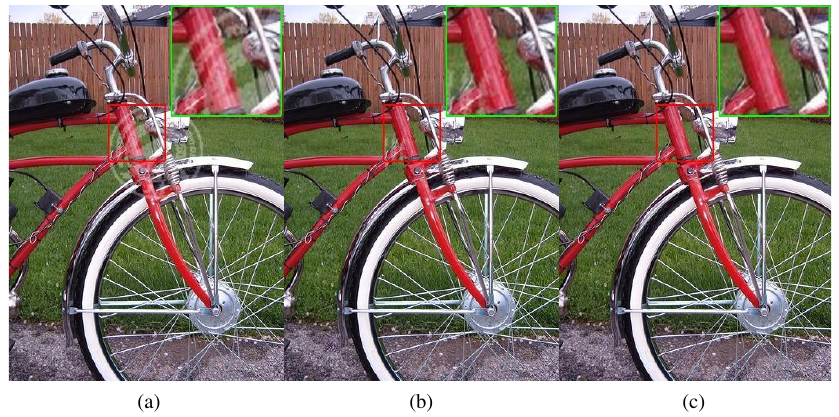}
  }
\caption{Visual images of different methods: (a) Watermarked image (29.21dB), (b) HN (32.18dB) and (c) SWCNN (35.75dB).}
 
\end{figure*}

\begin{table}[]
\centering
\caption{PSNR of different methods for different transparency.}
\begin{tabular}{lll}
\hline
Transparency               & Methods           & PSNR \\ \hline
\multirow{6}{*}{alpha=0.3} & Watermark   image & 28.9315 \\
                           & DnCNN  {\cite{zhang2017beyond}}            & 30.1071 \\
                           & FFDNet {\cite{zhang2018ffdnet}}            & 27.8820 \\
                           & DIP  {\cite{ulyanov2018deep}}               & 29.7473 \\
                           & WGAN-GP  {\cite{yu2018generative}}          & 31.0752 \\
                           &  {DRD-Net \cite{deng2019drd}}          & 28.9090\\
                           &  {FastDerainNet \cite{wang2020fastderainnet}}          & 32.2593\\
                           &  {EAFNWDD \cite{sun2021efficient}}          & 33.4744\\
                           & SWCNN(Ours) & 36.9022\\
\hline
\multirow{4}{*}{alpha=0.5} & Watermark   image & 24.4613 \\
                           & DnCNN  {\cite{zhang2017beyond}}            & 29.0817 \\
                           & FFDNet  {\cite{zhang2018ffdnet}}            & 26.6991 \\
                           &  {DRD-Net \cite{deng2019drd}}          & 29.5227 \\
                           &  {FastDerainNet \cite{wang2020fastderainnet}}          & 29.0564 \\
                           &  {EAFNWDD \cite{sun2021efficient}}          & 32.6197 \\
                           & SWCNN(Ours)  & 33.6491\\
\hline
\multirow{4}{*}{alpha=0.7} & Watermark image & 21.5422  \\
                           & DnCNN  {\cite{zhang2017beyond}}        & 27.8594 \\
                           & FFDNet  {\cite{zhang2018ffdnet}}         & 26.1591 \\
                           &  {DRD-Net \cite{deng2019drd}}          & 31.0825 \\
                           &  {FastDerainNet \cite{wang2020fastderainnet}}          & 25.5814 \\
                           &  {EAFNWDD \cite{sun2021efficient}}          & 27.0403\\
                           & SWCNN(Ours) & 30.8870\\
\hline
\multirow{4}{*}{alpha=1.0} & Watermark   image & 18.4946\\
                           & DnCNN   {\cite{zhang2017beyond}}          & 22.4960 \\
                           & FFDNet  {\cite{zhang2018ffdnet}}          & 20.4079 \\
                           &  {DRD-Net \cite{deng2019drd}}          & 23.6468 \\
                           &  {FastDerainNet \cite{wang2020fastderainnet}}          & 20.4333 \\
                           &  {EAFNWDD \cite{sun2021efficient}}          & 25.9116 \\
                           & SWCNN(Ours) & 28.1357\\
\hline
\end{tabular}
\end{table}

\begin{figure*}[!htbp]
\centering
\subfloat{\includegraphics[width=5.5in]{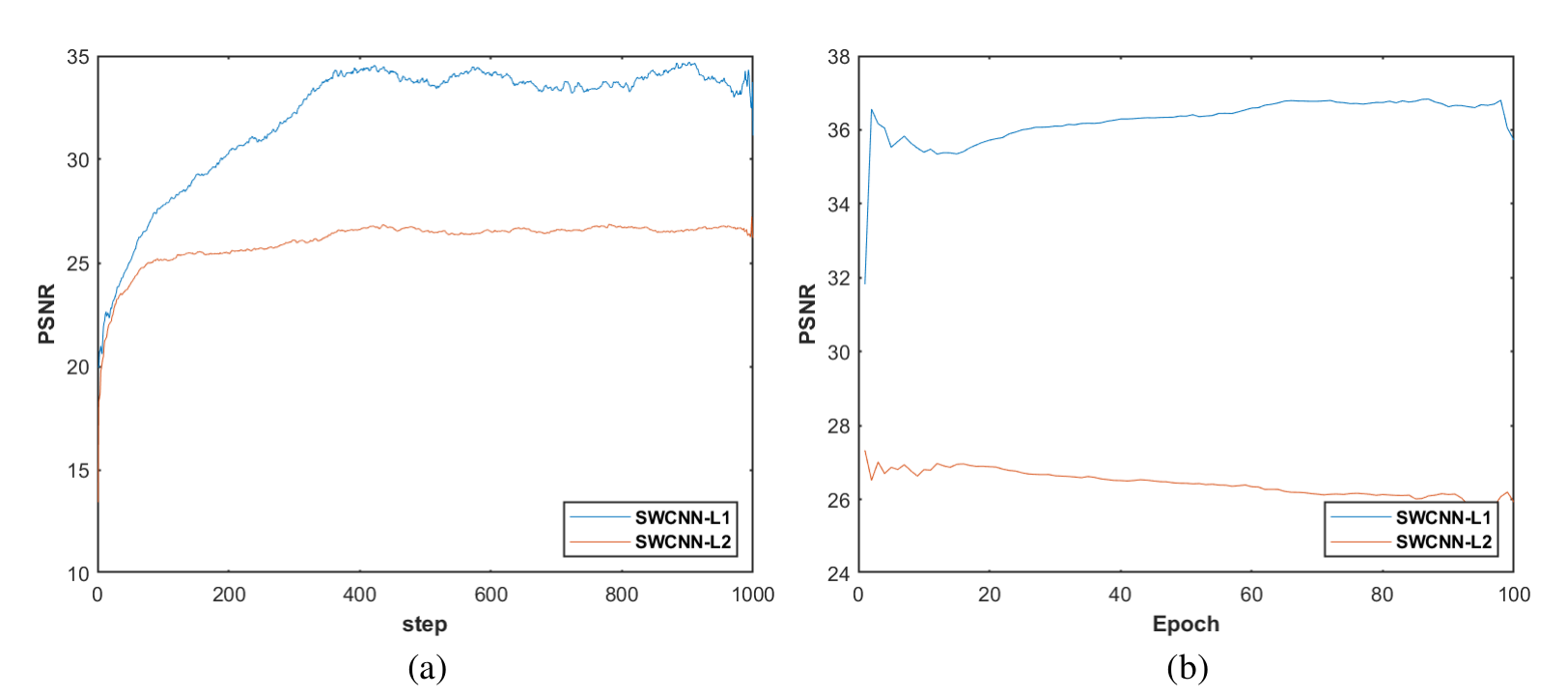}
  }
\caption{ {PSNR of different methods for one epoch and multiple epochs. (a) PSNR of two methods for one epoch, (b) PSNR of two methods for different epochs.}}
 
\end{figure*}

\begin{figure}[!htbp]
\centering
\subfloat{\includegraphics[width=3.5in]{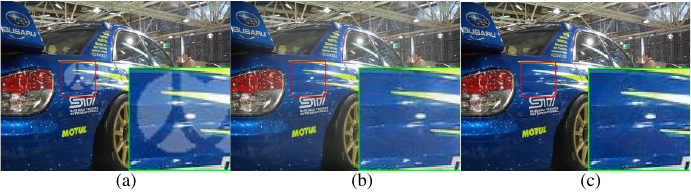}
  }
\caption{Visual images of different methods: (a) Watermarked image (29.82dB), (b) SWCNN with L2 (24.90dB), and (c) SWCNN with L1 (37.32dB).}
 
\end{figure}

\begin{figure}[!htbp]
\centering
\subfloat{\includegraphics[width=3.5in]{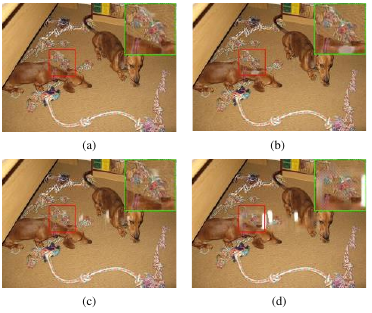}
  }
\caption{Visual effects of SWCNN for different transparency in image watermark. (a) SWCNN with transparency of 0.3 (40.72dB), (b) SWCNN with transparency of 0.5 (35.91dB), (c) SWCNN with transparency of 0.7 (33.65dB) and (d) SWCNN with transparency of 1 (26.05dB).}
 
\end{figure}

Mixed Loss: It is known that image representation, is very important for image restoration. Specifically, structural information from deep network \cite{sahu2018lightweight} and texture information from perception information \cite{li2019towards}. Inspired by that, we use a mixed loss to make a tradeoff between structural and texture information in this paper. That is, a mixed loss is composed of a structural loss and texture loss. The structural loss is conducted via obtained reference image and watermarked image from the proposed self-supervised mechanism in Section II.A. The texture loss is conducted via obtained features of middle layers of pretrained VGG to act ground truth and predicted image from the HN in Section II.D. 
  {The effectiveness of the mixed loss is verified via HN with clean reference images and SWCNN with clean reference images, HN and SWCNN in Table I. That is, a mixed loss above is more effective than a structural loss.} To verify robustness of the mixed loss, we monitor the training process for all the steps of one epoch and all the epochs in terms of PSNR.  {As shown in Fig.4, we can see that structural loss collaborates texture loss to achieve a mixed loss has better performance than that of a single structural loss for removing watermark.} Besides, we enlarge a chosen area of predicted image as an observation area of different methods (i.e., SWCNN and HN) to test performance of mixed loss for image watermark. As illustrated in Fig. 5, we can see that observation area of SWCNN is clear than that of HN, which shows effectiveness of texture loss.


\begin{figure*}[!htbp]
\centering
\subfloat{\includegraphics[width=5.5in]{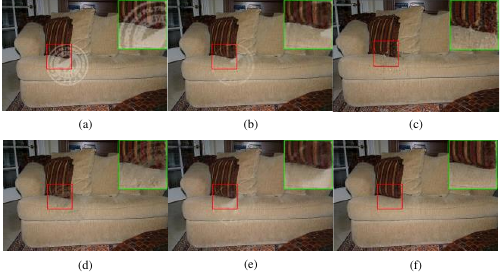}
  }
\caption{Visual effects of different methods with transparency of 0.3 for image watermark. (a) Watermarked image (29.82dB), (b) FFDNet (24.95dB), (c) DIP (34.88dB), (d) WGAN-GP (30.59dB), (e) DnCNN (28.26dB) and (f) SWCNN (Ours) (37.32dB).}
 
\end{figure*}


\subsection{Experimental results}

We choose several popular image restoration methods, i.e., DnCNN  \cite{zhang2017beyond}, FFDNet  \cite{zhang2018ffdnet}, DIP  \cite{ulyanov2018deep}, WGAN-GP  \cite{yu2018generative},  {DRD-Net\cite{deng2019drd}, FastDerainNet\cite{wang2020fastderainnet} and EAFNWDD\cite{sun2021efficient}} as comparative methods for quantitative and qualitative analysis. Quantitative analysis includes different methods for different transparency and same transparency, complexity (i.e., parameters and flops \cite{dolbeau2018theoretical}) and \textcolor{black}{image quality metrics, i.e., natural image quality evaluator (NIQE)\cite{mittal2012making} and Integrated Local NIQE (ILNIQE)\cite{zhang2015feature}}
to verify robustness of removing watermarks. For different transparency, we choose different transparency of 0.3, 0.5, 0.7 and 1 to act clean images to conduct experiments. As listed in Table III, we can see that the proposed SWCNN has obtained effective results for different transparency in image watermark. In terms of same transparency, transparency of 0.3 is chosen to conduct experiments. As reported in Table II, we can see that our SWCNN has obtained the highest results in terms of PSNR, which shows effectiveness of SWCNN with certain transparency for image watermark. To evaluate computational cost of SWCNN, we choose complexity to conduct experiments. As descripted in Table IV, we can see that the proposed method is more competitive than other popular methods, i.e. WGAN-GP and DRD-Net in parameters and flops. \textcolor{black}{To prevent limitations of PSNR and SSIM for low-level vision tasks, image quality evaluators are used to conduct comparative experiments in this paper. Taking into blind image watermark removal account, popular blind image quality evaluators \cite{wu2015highly, wu2015blind} are prioritized to measure obtained non-watermark images. Li et al. \cite{wu2015highly} utilized binary patterns of local image structures to extract statistical information to achieve blind image quality assessment. Alternatively, a fusions of statistical information based multiple domains and different channels was presented to address image quality problem \cite{wu2015blind}. To fairly test quality of obtained images from our SWCNN, we choose blind image quality evaluators with popular codes, i.e., NIQE\cite{mittal2012making} and (ILNIQE)\cite{zhang2015feature} to test comparative experiments in TABLE VI. As shown in TABLE VI, we can see that obtained SWCNN has obtained the lowest value on both NIQE and ILNIQE than that of other popular methods, i.e., DnCNN, FFDNet and FastDerainNet, which shows it is very competitive for  blind image quality assessment.}  According to mentioned illustrations, the proposed SWCNN is effective in image watermark in terms of quantitative analysis.


\begin{table}[]
\centering
\caption{Complexity of different methods on an image with $256\times256$  for image watermark.}
\begin{tabular}{cccc}
\hline
Methods          &  & Parameters & Flops   \\ \hline
FFDNet \cite{zhang2018ffdnet}  &  & 854.688K   & 14.003G \\
DIP \cite{ulyanov2018deep}     &  & 2.996M     & 1.776G  \\
WGAN-GP \cite{yu2018generative} &  & 3.602M     & 22.470G \\
DnCNN  \cite{zhang2017beyond}   &  & 558.336K   & 36.591G \\
 {DRD-Net \cite{deng2019drd}} &  &  {2.941M}     &   {192.487G} \\
 {FastDerainNet \cite{wang2020fastderainnet}} &  &  {336.006K}     &  {22.128G} \\
 {EAFNWDD \cite{sun2021efficient}}         &  &  {52.322M}     &  {135.007G}\\
SWCNN (Ours)     &  & 700.611K   & 18.624G \\ \hline
\end{tabular}
\end{table}

\begin{table}[]
\centering
\caption{PSNR and SSIM of SWCNN with different losses.}
\begin{tabular}{ccc}
\hline
Methods  & PSNR    & SSIM   \\ \hline
SWCNN-L1 & 36.9022 & 0.9893 \\
SWCNN-L2 & 25.3875 & 0.9383 \\ \hline
\end{tabular}
\end{table}

\begin{table}[]\color{black}
\centering
\caption{NIQE and ILNIQE of different methods for transparency of 0.3.}
\begin{tabular}{lll}
\hline
            Methods & NIQE\cite{mittal2012making} & ILNIQE\cite{zhang2015feature} \\ \hline 
                            DnCNN  {\cite{zhang2017beyond}}  & 4.8500 &23.2775\\
                            FFDNet {\cite{zhang2018ffdnet}} & 5.5887 &25.3639\\
                             {FastDerainNet \cite{wang2020fastderainnet}} &4.8070 & 22.9647\\
                            SWCNN(Ours) & 4.8039 &22.9337\\
\hline
\end{tabular}
\end{table}

\begin{figure}[!htbp]
\centering
\subfloat{\includegraphics[width=3.5in]{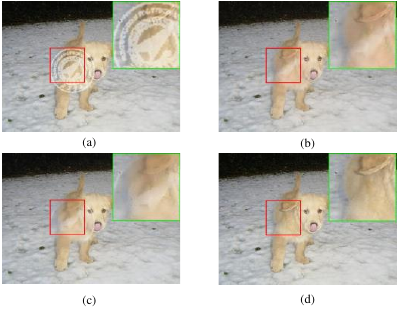}
  }
\caption{Visual effects of different methods with transparency of 0.5 for image watermark. (a) Watermarked image (34.24dB), (b) DnCNN (33.46dB), (c) FFDNet (31.50dB) and (d) SWCNN (Ours) (40.89dB).}
 
\end{figure}

To accelerate training speed, L1 rather than L2 is embedded into structural and texture loss. We use quantitative and qualitative analysis to evaluate performance SWCNN with L1 and SWCNN with L2. In terms of quantitative analysis, average PSNR and SSIM of SWCNN with L1 and SWCNN with L2 on all the test images are used to conduct experiments. As shown in Table V, we can see that SWCNN with L1 is higher PSRN and SSIM than that of SWCNN with L2, which shows effectiveness of loss function with L1 for image watermark. On the other hand, we monitor the training process for all the steps of one epoch and all the epochs in terms of PSNR to verify effectiveness of loss function with L1. As presented in Fig. 6, we can see that SWCNN with L1 has obtained higher values than these of SWCNN with L2 for all the steps of one epoch and multiple epochs, which shows effectiveness of loss function with L1 for the whole training process in image watermark. In terms of qualitative analysis, we can see that observation area of SWCNN with L1 is clear than that of SWCNN with L2, which shows effectivness of loss function with L1 as shown in Fig. 7. According to quantitative and qualitative analysis, loss function with L1 is suitable to SWCNN for image watermark removal.

Qualitative analysis: we amplify an area of different methods as observation area to evaluate visual effects via their clarity. Fig. 8 shows observation areas of our SWCNN for different transparency, which shows our method is robust for different transparency. Fig. 9 and Fig. 10 show that the proposed SWCNN is clearer than other popular methods, i.e., FFDNet and DnCNN for removing watermarks. That shows the proposed SWCNN is effective for image watermark removal in terms of qualitative analysis. According to quantitative analysis and qualitative analysis, we can see that the proposed SWCNN is suitable to image watermark removal . 

\section{Conclusion}
In this paper, we propose a self-supervised convolutional neural network for image watermark removal as well as SWCNN. Proposed SWCNN uses a self-supervised way to construct reference images rather than given paired training samples, according to watermark distribution. A heterogenous U-Net and a mixed loss are used to make a tradeoff between structural and texture information. Besides, a watermark dataset with twelve different watermarks for different coverages is conducted to verify robustness of the proposed watermark method. Our SWCNN is competitive compared with popular CNNs in terms of different transparency, visual effects, complexities for image watermark removal. Evaluations on widely used benchmarks demonstrate the effectiveness of our proposed method.





\ifCLASSOPTIONcaptionsoff
  \newpage
\fi




%

\bibliographystyle{IEEEtran}
\bibliography{IMSC_AGL}

\end{document}